# AN OPTIMIZATION METHOD FOR SLICE INTERPOLATION OF MEDICAL IMAGES

*Ahmadreza Baghaie[1], Zeyun Yu[1]*

[1]Department of Electrical Engineering and Computer Science, University of Wisconsin-Milwaukee, WI, USA

**ABSTRACT**

Slice interpolation is a fast growing field in medical image processing. Intensity-based interpolation and object-based interpolation are two major groups of methods in the literature. In this paper, we describe an object-oriented, optimization method based on a modified version of curvature-based image registration, in which a displacement field is computed for the missing slice between two known slices and used to interpolate the intensities of the missing slice. The proposed approach is evaluated quantitatively by using the Mean Squared Difference (MSD) as a metric. The produced results also show visual improvement in preserving sharp edges in images.

## 1. INTRODUCTION

Image interpolation is a well-known research topic in image processing and there have been many studies in this area especially in bio-medical applications. With modern image modalities (CT, MRI, light/electron microscopy, etc.), a sequence of 2D images can be provided and used in building 3D models. However, the resolutions of the images are often not identical in all three directions. Usually the resolution in the Z direction is significantly lower than the resolutions in the X and Y directions. For example, in a generic CT, resolution in X and Y direction, or in more accurate term the spacing, is between 0.5-2mm while the spacing in Z direction is in the range of 1-15mm. This asymmetry in the resolution causes problems such as step-shaped iso-surfaces and discontinuity in structures in 3D reconstructed models. Therefore utilizing a slice interpolation algorithm to augment the 3D data into a symmetric one is of high demand.

In general, slice interpolation methods can be divided into two groups: intensity-based interpolation, and object-based interpolation. In the first category, the final result of interpolation is directly computed from the intensity values of input images. Linear and cubic spline interpolation methods are two examples of this group. The major advantages of these methods are their simplicity and low computational complexity, which lead to their wide uses in practice. As the final result is basically a weighted average of input image, however, these methods suffer from blurring effects on object boundaries, yielding unrealistic and visually unpleasing results.

In object-based methods, on the other hand, the extracted information from objects contained in input images are used in order to guide the interpolation into more accurate results. There are many methods proposed in the literature trying to take into account additional information of objects in order to provide better results [1-11]. One of the first attempts for object-based interpolation has been made by Goshtasby et al [1]. Using a gradient magnitude based approach, corresponding points between consecutive slices are found and then the linear interpolation is applied in order to find the in-between slices. An important assumption of this work is that the difference between consecutive slices is small, so they restrict their search for finding correspondence points to small neighborhoods. It is obvious that this assumption is not true in many cases. To reduce the blurriness of edges, some more recent approaches have been studied, including the *column fitting interpolation* [2], *shape-based method* [3], *morphology-based method* [6], and *feature-guided shape interpolation method* [7]. A comprehensive summary of common methods (both intensity-based and object-based) for slice interpolation was described in [4, 5].

An increasingly important group of approaches for image interpolation (object-based) are based on image registration. Using the well-known free form deformation non-rigid registration method by Rueckert [8], Penny et al. [9] proposed a registration based method for slice interpolation. Another registration based method was given by Frakes et al. [10] by using a modified version of *control grid interpolation* (CGI). More recently, Xu et al. [11] described a multi-resolution registration based method for slice interpolation. In general, registration-based slice interpolation methods are guided by two important assumptions. First, the consecutive slices contain similar anatomical features. Second, the registration method is capable of finding the appropriate transformation map to match these similar features. Violation of any of these assumptions results in false correspondence maps, which leads to incorrect interpolation results.

In the present paper, we develop a novel method for slice interpolation by taking into account the well-known curvature-based registration [12, 13]. With a modified version of the registration method and an assumption of having linear movement between correspondence points in given slices, a displacement field is computed and the in-between slice is interpolated using a simple averaging of the registration results. The detail of the proposed method is given in Section 2, followed by some experimental results along with quantitative and qualitative evaluations of the method in Section 3. The conclusion is given in Section 4.

## 2. METHOD

Given two images (*reference* and *template*), image registration is to find a spatial transformation such that the transformed template matches the reference, subject to a suitable distance measure (forward registration) [12]. This transformation can range from a simple translation to more sophisticated non-rigid free form deformations, dependent on the subject and goal of registration. Also depending on the method, registration can be based on matching a set of feature points (landmarks) or been applied directly on image gray values. Here we are interested in the later case. Usually most of the registration methods can be formulated in term of a variational formulation, using a joint functional as follows [12]:

$$E(\boldsymbol{u}) = D[R, T; \boldsymbol{u}] + \alpha S[\boldsymbol{u}], \qquad (1)$$

where $D$ represents a distance measure (external force) and $S$ represents the rate of smoothness of $\boldsymbol{u}$ (internal force). The parameter $\alpha$ is used to balance the two terms. In this functional, $\boldsymbol{u}$ should be found such that the joint functional is minimized. This model is called *single direction model* because the reference image is fixed and only the template image is moving. This causes asymmetry in the results in such a way that if we fix the template image and move the reference image to match the template image (backward registration), the result may not be exactly opposite to that of the forward registration. For this reason, we modify this model in the context of image slice interpolation by changing the formulation to the following:

$$E[\boldsymbol{u}] = D[R_1(\boldsymbol{x} - \boldsymbol{u}), R_2(\boldsymbol{x} + \boldsymbol{u})] + \alpha S[\boldsymbol{u}], \qquad (2)$$

where $R_1, R_2 : \Omega \to \mathbb{R}$ are the two images provided as input and $\Omega \coloneqq [0,1]^2$ is the domain of images, $\boldsymbol{x}$ is the grid of image values and $\boldsymbol{u}$ is the displacement values for each grid point. Please note that in Equation (2), we assume that the slice to be interpolated, denoted by $R$, is in the middle of the given images. If $R$ is an arbitrary slice between $R_1$ and $R_2$, then we first need to compute the distance from $R$ to $R_1$ and $R_2$, denoted by $d_1$ and $d_2$ respectively. Then we calculate the ration $r = d_1/(d_1+d_2)$, and the following equation should be considered for interpolating $R$:

$$E[\boldsymbol{u}] = D[R_1(\boldsymbol{x} - r\boldsymbol{u}), R_2(\boldsymbol{x} + (1-r)\boldsymbol{u})] + \alpha S[\boldsymbol{u}]. \qquad (3)$$

Without loss of generality, we shall consider Equation (2) in the current paper for image slice interpolation.

Several distance measures for D have been proposed in the literature, including the Sum of Squared Differences (SSD), Mutual Information (MI), Normalized Mutual Information (NMI), Cross Correlation (CC) and Normalized Gradient Fields (NGF) [14]. Here we use SSD as distance measure, and the above formulation can rewritten as:

$$D[R_1(\boldsymbol{x}-\boldsymbol{u}), R_2(\boldsymbol{x}+\boldsymbol{u})] = \tfrac{1}{2}|R_1(\boldsymbol{x}-\boldsymbol{u}) - R_2(\boldsymbol{x}+\boldsymbol{u})|_{L_2}^2 = \tfrac{1}{2}\int_\Omega (R_1(\boldsymbol{x}-\boldsymbol{u}(\boldsymbol{x})) - R_2(\boldsymbol{x}+\boldsymbol{u}(\boldsymbol{x})))^2 d\boldsymbol{x}. \qquad (4)$$

For the smoothness term $S$, several common choices are available, such as elastic, fluid, demon, diffusion and curvature registration [12]. Here we use the curvature approach, in which the smoothness term is as follows:

$$S[u] = \tfrac{1}{2}\sum_{l=1}^{2}\int_\Omega (\Delta u_l)^2 d\boldsymbol{x}, \qquad (5)$$

where $\Delta$ is the curvature operator and the summation is computed over two dimensions of image and the integral is computed inside the domain of images. As stated in [12], using curvature for smoothness, the need for an additional linear affine pre-registration step can be eliminated.

In order to minimize the above joint functional, we compute the Gateaux derivative of $E[\boldsymbol{u}]$ and make it equal to zero to find the minimum point, an Euler-Lagrange PDE equation can be obtained as:

$$f(\boldsymbol{x}, \boldsymbol{u}(\boldsymbol{x})) + \alpha \mathcal{A}^{curv}[\boldsymbol{u}](\boldsymbol{x}) = 0, \qquad (6)$$

where

$$\mathcal{A}^{curv}[\boldsymbol{u}] = \Delta^2 \boldsymbol{u} \qquad \text{and}$$

$$f(\boldsymbol{x}, \boldsymbol{u}(\boldsymbol{x})) = (R_2(\boldsymbol{x}+\boldsymbol{u}) - R_1(\boldsymbol{x}-\boldsymbol{u})).(\nabla R_1(\boldsymbol{x}-\boldsymbol{u}) + \nabla R_2(\boldsymbol{x}+\boldsymbol{u}))$$

To solve this PDE, a time-stepping iteration method is considered as follows:

$$\partial_t \boldsymbol{u}^{k+1}(\boldsymbol{x},t) = f\left(\boldsymbol{x}, \boldsymbol{u}^k(\boldsymbol{x},t)\right) + \alpha \mathcal{A}^{curv}[\boldsymbol{u}^{k+1}](\boldsymbol{x},t), \quad k \geq 0 \quad (7)$$

with $\boldsymbol{u}^0 = \boldsymbol{0}$. Using a finite difference approximation of the derivative with time step $\tau$ and also collecting the grid points with respect to a lexicographical ordering, one can derive a discretized version of (7) as follows:

$$(I_n + \alpha\tau\mathcal{A}^{curv})\overline{\boldsymbol{U}}_l^{(k+1)} = \overline{\boldsymbol{U}}_l^{(k)} + \tau\overline{\boldsymbol{F}}_l^{(k)}, \quad l = 1,2. \quad (8)$$

where $l$ is the parameter representing the dimension index. For more information about how to compute $\mathcal{A}^{curv}$, the reader is referred to [12, 13]. After finishing the optimization process, a simple averaging of the two transformed input images provides us with the missing in-between image.

## 3. RESULTS AND DISCUSSION

To validate the proposed method for slice interpolation in medical images, several tests have been conducted. The results of the proposed method are compared with two other methods, in both subjective and objective aspects. As a metric, Mean Squared Difference (MSD) is used for comparison. Assuming $I_{org}$ and $I_{int}$ as original image and interpolation image respectively, with the size of $m \times n$, we define MSD as follows:

$$MSD = \frac{1}{m \times n}\sum_{i=1}^{m}\sum_{j=1}^{n}(I_{org}(i,j) - I_{int}(i,j))^2 \quad (9)$$

In the first test, three consecutive slices as in Fig. 1(a) are used. Taking the first and third slices as inputs, we compute the in-between slice by using both linear interpolation and the proposed method. Fig. 1 (b) shows the interpolation results (top) using the two methods as well as the computed difference images (bottom) with respect to the second image in Fig. 1(a). The MSDs computed are 84.20 and 54.96 for linear and proposed methods respectively. Fig.1 (c) gives the displacement fields computed after registration, for both horizontal and vertical directions. Bright and dark shades represent positive and negative displacement vectors respectively, while gray shades for displacements near zero.

As can be seen from the slices, the difference between slices is due to the movement of the heart near the centers of these images. Using linear interpolation, the movement of heart is not captured, resulting in blurred edges in the interpolated slice. By comparison, the registration-based method captures the movement well and the final result is highly similar to the original one (middle image in Fig. 1(a)). As a result, the MSD error is significantly reduced and the interpolation result is much sharper.

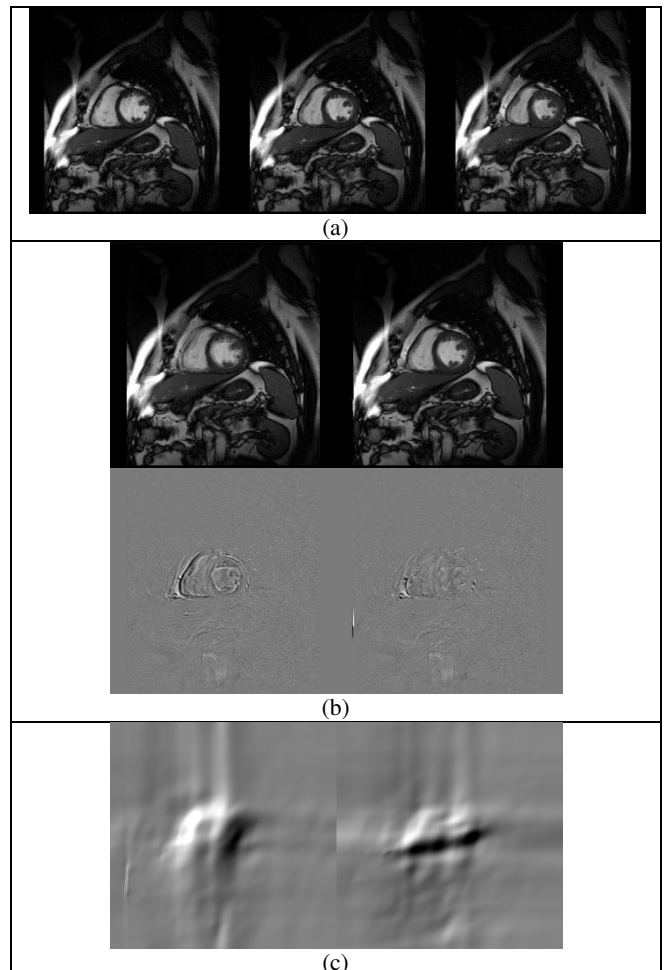

Fig. 1 (a) Three consecutive images. The first and third images are used for interpolation. To produce the results the parameters values are set as $\tau = 0.03, \alpha = 100$. (b) Top row: interpolation results for linear and proposed method. Bottom row: difference images of the results and reference image (middle image in (a)). (c) Displacement fields for horizontal and vertical directions using the proposed method.

To further demonstrate the strength of the proposed method, we apply the same procedure to the second example containing three brain images as shown in Fig. 2 (a). Using the first and third slice, the interpolation results are produced. Besides the linear interpolation, we also compare the proposed method with the original curvature registration based technique [12], called *non-modified method* below. Fig. 2(b) shows the results of interpolation as well as the computed difference images. The MSDs are 71.65, 45.36 and 42.72 for linear, non-modified and proposed method respectively. As can be seen, the result of linear interpolation has uncertain and highly blurred edges. Result of non-modified method is significantly better than linear interpolation, in terms of MSD but due to nonlinear nature of image registration and optimization process, we still have blurred edges. In comparison, the proposed method gives much sharper edges. This becomes more obvious in the difference images where blurred edges yield widened regions of dissimilarity. Also, it should be mentioned that, in the non-modified method, only one of the images is moving. As a result, more iterations of optimization are needed for convergence, and thus more computational time is required.

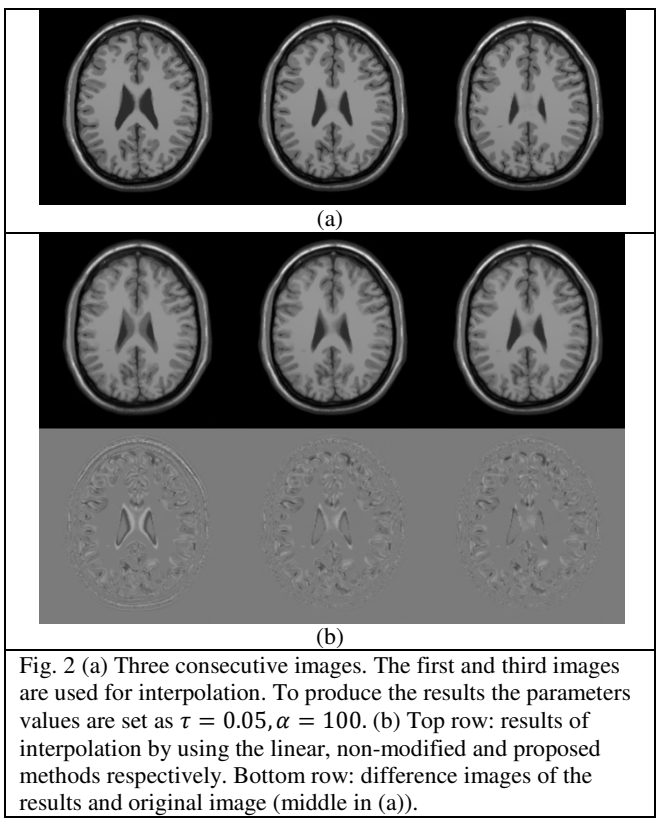

Fig. 2 (a) Three consecutive images. The first and third images are used for interpolation. To produce the results the parameters values are set as $\tau = 0.05, \alpha = 100$. (b) Top row: results of interpolation by using the linear, non-modified and proposed methods respectively. Bottom row: difference images of the results and original image (middle in (a)).

## 4. CONCLUSION

In this paper, a new registration-based slice interpolation method is proposed. A modified version of curvature registration method has been used with the assumption of linear displacements between corresponding points in two input images. The obtained displacement fields for the two images are utilized to produce the missing in-between slice. In comparison to both linear interpolation and the non-modified registration-based method [12], the proposed method produces lower MSD values and less blurred/uncertain edges. The current implementation was performed in Matlab, which takes a total of ~30 seconds for an image of 250*250 pixels. An ongoing effort is to implement the algorithm in C/C++, which is expected to consume much less computational time. Part of our future work will be an extension of the linear displacements between images into higher order polynomial formulation that involves more than two adjacent slices.